\documentclass[lettersize,journal]{IEEEtran}
\usepackage{amsmath,amsfonts}
\usepackage{algorithmic}
\usepackage{algorithm}
\usepackage{array}
\usepackage[caption=false,font=normalsize,labelfont=sf,textfont=sf]{subfig}
\usepackage{textcomp}
\usepackage{stfloats}
\usepackage{url}
\usepackage{verbatim}
\usepackage{graphicx}
\usepackage{cite}
\usepackage[colorlinks=true, citecolor=blue, linkcolor=red]{hyperref}
\usepackage{multirow}

\usepackage{orcidlink}

\begin{document}

\title{RESAR-BEV: An Explainable Progressive Residual Autoregressive Approach for Camera-radar Fusion in BEV Segmentation}

\author{Zhiwen Zeng\orcidlink{0009-0009-3391-4080}, Yunfei Yin\orcidlink{0000-0003-1792-1378},~\IEEEmembership{Member, IEEE}, Zheng Yuan\orcidlink{0009-0005-8454-4383}, Argho Dey\orcidlink{0009-0000-6739-9632}, and Xianjian Bao\orcidlink{0000-0001-7197-6738}

\thanks{Manuscript received 9 May 2025; revised 5 October 2025 and 26 January 2026; accepted 1 March 2026. This research work has been partially supported by the National Natural Science Foundation of China (62262045), the Fundamental Research Funds for the Central Universities (2023CDJYGRH-YB11), and the Open Funding of SUGON Industrial Control and Security Center (CUIT-SICSC-2025-03). (Corresponding author: Yunfei Yin.)}

\thanks{Zhiwen Zeng, Yunfei Yin, Zheng Yuan, and Argho Dey are with the College of Computer Science, Chongqing University, Chongqing 400044, China (e-mail: zwzengi@outlook.com; yinyunfei@cqu.edu.cn; yuanzheng@cqu.edu.cn; arghomridul05@gmail.com).}

\thanks{Xianjian Bao is with the Department of Computer Science, Maharishi University of Management, Fairfield, IOWA, USA (e-mail: xibao@mum.edu).}
}

\IEEEpubid{0000--0000/00\$00.00~\copyright~YYYY IEEE}

\markboth{IEEE TRANSACTIONS ON INTELLIGENT TRANSPORTATION SYSTEMS,~Vol.~xx, No.~xx, Month~YYYY}%
{Shell \MakeLowercase{\textit{et al.}}: A Sample Article Using IEEEtran.cls for IEEE Journals}

\maketitle

\begin {abstract}
Bird's-Eye-View (BEV) semantic segmentation provides comprehensive environmental perception for autonomous driving, but suffers from multimodal misalignment and sensor noise. We propose RESAR-BEV, a progressive refinement framework that advances beyond single-step end-to-end approaches:
(1) an inherently interpretable, coarse-to-fine refinement using a residual autoregressive learning paradigm implemented by our Drive-Transformer and Modifier-Transformer cascade, where each stage is responsible for a specific semantic scale (from road topology to lane boundaries);  (2) robust BEV representation combining ground-proximity voxels with adaptive height offsets and dual-path voxel feature encoding (max+attention pooling) for efficient feature extraction; and (3) decoupled supervision with offline Ground Truth decomposition and online joint optimization, which prevent overfitting while ensuring structural coherence.
Experiments on nuScenes demonstrate RESAR-BEV achieves state-of-the-art performance with 54.0\% mIoU across 7 essential driving-scene categories while maintaining real-time capability at 14.6 FPS. The framework exhibits robustness in challenging scenarios of long-range perception and adverse weather conditions. 
\end {abstract}


\section{Introduction}
\IEEEPARstart{A}{utonomous}  driving systems require comprehensive 3D environment understanding to ensure safe navigation. While conventional perception tasks (e.g., object detection~\cite{qi2018frustum,wang2021fcos3d,kim2019deep,craft,centerfusion}, semantic segmentation~\cite{lawin2017deep,boulch2017unstructured,guerry2017snapnet}) operate within camera frustum views, Bird's Eye View (BEV) representation has emerged as a pivotal paradigm for unifying multi-sensor inputs (including cameras, radars, and LiDARs) into a cohesive 3D scene representation.

\begin{figure}[t]
    \centering
    \includegraphics[width=\linewidth]{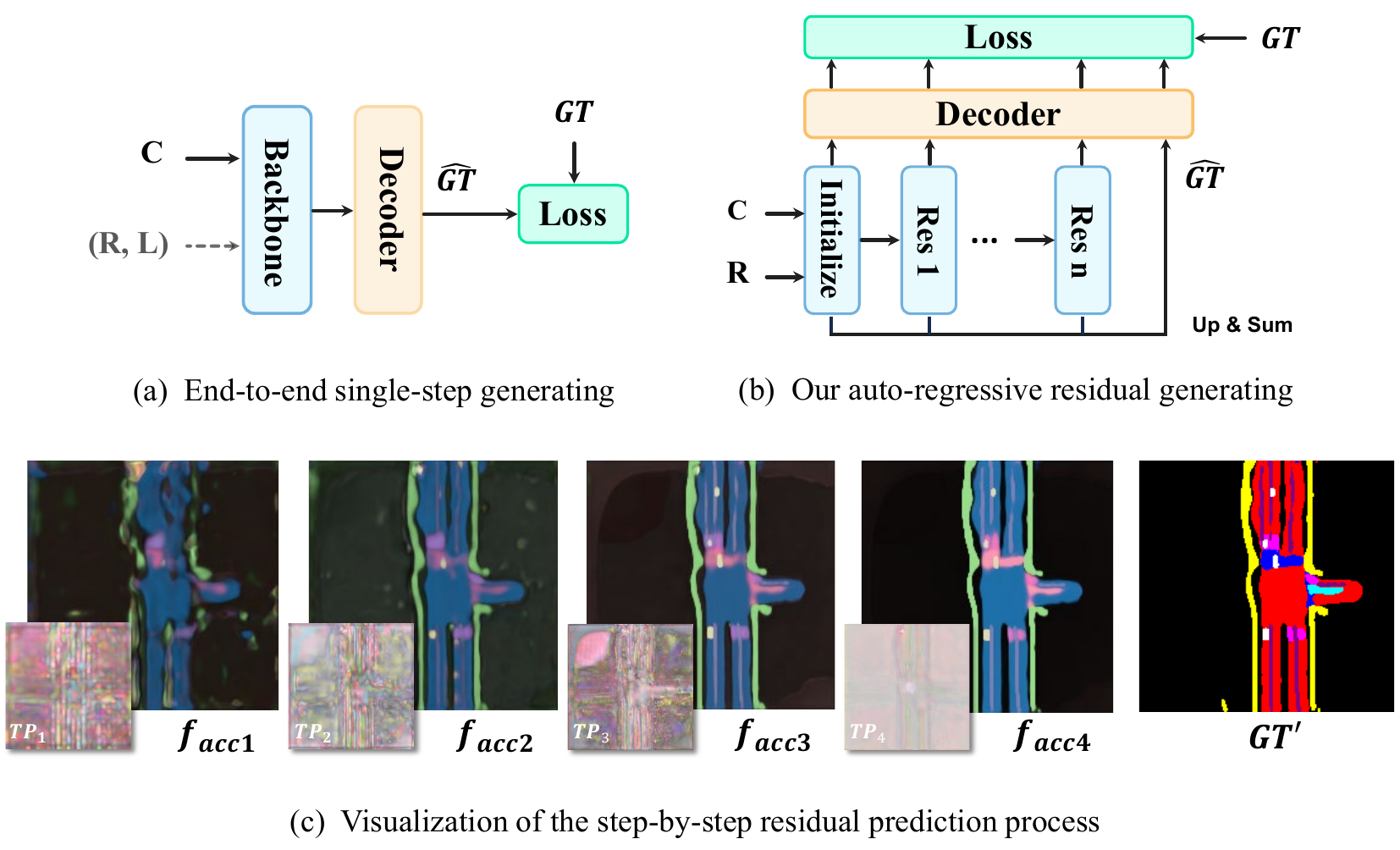}
    \vspace{-.6cm}
    \caption{(a) Previous single-step end-to-end multimodal BEV scene segmentation approach; (b) Our proposed progressive autoregressive residual prediction method with multi-resolution Ground Truth and multi-stage loss supervision; (c) Visualization of the Residual Autoregressive Fusion module's multi-scale feature generation and accumulation process through decoder mapping to BEV semantics, followed by upsampling to uniform resolution.}
    \label{fig:intro}
\end{figure}

BEV segmentation partitions BEV space into semantic regions like drivable areas and vehicles. Current segmentation approaches based on BEV representation follow two paradigms: geometry-based methods (e.g., IPM~\cite{ipm}) requiring precise calibration but lacking robustness, and learning-based methods (e.g., BEVFormer~\cite{bevformer}, Bev-locator~\cite{bevlocator}) automatically learning cross-modal correlations but needing careful design to address noise and misalignment. While recent works also adopt multimodal fusion and temporal smoothing ~\cite{polarbev,polarformer,bevdet4d,beverse,retentivebev}, they employ an end-to-end paradigm, which forces the network to generate the final BEV layout in a single step (Fig.\ref{fig:intro} (a)), neglecting the hierarchical spatial reasoning process from road topology to lane-level details. The lack of effective supervision over the intermediate stages from multimodal inputs to the final output makes the system vulnerable to errors in depth estimation or cross-modal alignment, the impact of which would be global and difficult to trace.

In modality selection, camera-radar fusion leverages their complementary advantages for BEV segmentation. Cameras deliver rich semantics but are vulnerable to environmental variations, whereas radar provides robust spatial perception in adverse weather and is superior to LiDAR in cost and latency for real-time obstacle avoidance. Despite its sparsity, radar compensates for camera weaknesses by providing critical depth and structural information, particularly for long-range and low-visibility cases. Thus, integrating both modalities results in a significantly more robust and practical system.

\IEEEpubidadjcol

Inspired by the progressive nature of human driving cognition and by methodologies in residual learning, we present RESAR-BEV, a framework that reformulates BEV segmentation as a process of progressive residual refinement. As illustrated in Fig.\ref{fig:intro} (b), RESAR-BEV decomposes the task into: (1) coarse low-resolution BEV initialization, (2) residual cascade refinement with progressive residual accumulation, (3) a hierarchical loss supervision enforcing coarse topological to fine pixel-level consistency.

Evaluated on the nuScenes dataset, RESAR-BEV exhibits the anticipated progressive residual prediction behavior (Fig. \ref{fig:intro} (c)): early stages establish the coarse road layout, while later stages add high-frequency details such as vehicles and lane dividers, demonstrating a semantically hierarchical and interpretable process. Our main contributions are as follows:

(1) \textbf{Progressive Residual Autoregressive Learning}: We introduce a novel paradigm that decomposes BEV segmentation into a coarse-to-fine process via a cascaded Transformer, with multi-scale ground truth decomposition and dynamic gating to stabilize training and enable error localization.

(2) \textbf{Ground-Aware BEV Optimization}: We introduce ground-proximity voxels with adaptive height offsets, enhancing BEV spatial modeling through our improved dual-path radar encoding (max+attention pooling) with intermediate feature fusion to boost long-range and low-light robustness at minimal computational overhead.

(3) \textbf{Decoupling Supervision and Prediction}: We employ offline pre-training of a GT decomposition network with online joint optimization of residual and segmentation losses to mitigate overfitting. Early stages capture global structures and later stages refine local details, mimicking human driving cognition and providing inherent support for visual interpretability.

\section{Related Work}

BEV representation is fundamental for 3D object detection~\cite{geobev,bevcam3d} and HD map segmentation~\cite{bevcar,compbev}, yet challenges persist in cross-modal alignment, progressive refinement, and model interpretability. In this section, we analyze these gaps through three key dimensions.

\subsection{BEV Segmentation Paradigms}
Early BEV segmentation adopted implicit geometric modeling, learning image-to-BEV mappings directly through neural networks (VED~\cite{ved} with VAEs, VPN~\cite{vpn} with MLPs). However, their purely data-driven nature without geometric priors constrained performance in complex scenarios.

Explicit geometric modeling has become mainstream by incorporating geometric priors for improved robustness. Early approaches primarily used Inverse Perspective Mapping (IPM~\cite{ipm}), which transforms camera views to BEV via geometric projection. While relying on idealized assumptions (e.g., flat ground, precise calibration), IPM established key geometric constraints that guided subsequent research.

The Lifting \& Unlifting paradigm significantly advanced explicit 3D modeling for view transformation. Lifting methods (Image$\rightarrow$3D$\rightarrow$BEV) first project image features into 3D space before BEV mapping. LSS~\cite{lss} pioneered this approach through soft depth distribution estimation, while BEVDet~\cite{bevdet} optimized feature extraction and FIERY~\cite{fiery} enhanced dynamic understanding via temporal fusion. BEVCar~\cite{bevcar} employed direct voxel projection, preserving features at higher computational cost. Conversely, Unlifting methods (BEV Query$\rightarrow$Image) use BEV queries for reverse projection. DETR3D~\cite{detr3d} introduced Transformer-based queries, improved by Deformable DETR's~\cite{deformabledetr} learnable sampling. BEVFormer~\cite{bevformer} enhanced cross-view attention with temporal fusion, and BEVSegFormer~\cite{bevsegformer} adapted queries specifically for segmentation tasks. These approaches balanced accuracy and efficiency in feature transformation.

The paradigm has shifted from purely data-driven implicit methods to geometry-aware hybrid approaches, optimizing the accuracy-efficiency-interpretability trade-off in BEV perception. Following this latest paradigm, our work leverages "unlifting" to enable interactions between BEV queries and multi-scale image features.

\subsection{Multi-Model Fusion for BEV Perception}
Multimodal fusion has become crucial for robust BEV segmentation, overcoming single-sensor (e.g., camera) limitations in challenging conditions (e.g., night, rain) through two main approaches: camera-LiDAR and camera-radar fusion.

In camera-LiDAR fusion, BEVFusion~\cite{bevfusion} employs feature concatenation followed by a convolution-based encoder for efficient intermediate fusion, while TransFusion~\cite{transfusion} leverages Transformer for finer cross-modal feature interaction. For cost-effective radar sensors, researchers have proposed innovative approaches: Simple-BEV~\cite{simplebev} achieves rapid fusion via grid-based processing, FISHING~\cite{fishing} introduces category-first pooling, CRN~\cite{crn} incorporates deformable attention mechanisms, and BEVGuide~\cite{bevguide} proposes a unified BEV space query method. These advancements collectively drive the evolution of fusion strategies from early-stage simple feature concatenation to attention-based adaptive fusion.

Recent works have introduced more innovative approaches. BEVFormer v2~\cite{bevformerv2} introduces perspective supervision to mitigate the inherent optimization difficulties when adapting a generic image backbone for BEV modeling; StreamPETR~\cite{streampetr} enhances dynamic video stream detection by propagating object queries across time; RecurrentBEV~\cite{recurrentbev} proposes a long-sequence fusion framework for object detection; and DualBEV~\cite{dualbev} introduces a unified 3D-to-2D feature transformation approach. Bev-tsr~\cite{bevtsr} presents a BEV-text retrieval system using LLM-processed descriptions and knowledge graph embeddings. Our method implements early dual-branch feature extraction: an vision extraction network and optimized voxel feature encoder, followed by intermediate cross-modal feature fusion.

\subsection{Autoregressive Residual Learning and Interpretability}
While progressive residual learning has demonstrated success in image segmentation (e.g., RMS-UNet~\cite{RMS-UNet}) and autoregressive image generation (e.g., VAR~\cite{var}, RQ-VAE~\cite{rqvae}), its potential in BEV perception remains largely unexplored. Current BEV segmentation methods predominantly adopt a single-step end-to-end global prediction paradigm (e.g., CVT~\cite{cvt}, SparseFusion3D~\cite{sparsefusion3d}), which lacks explicit supervision during the generation process. This monolithic approach suffers from irreversible error accumulation and behavioral misalignment with human driving cognition, which progressively refines scene understanding from coarse road topology to fine lane boundaries—a process fundamentally mismatched with parallel, single-step decoding paradigms. Inspired by this observation, our model decomposes the task of constructing the entire BEV segmentation map into relatively independent residuals and achieves a coarse-to-fine generation process through an autoregressive mechanism.

\begin{figure*}[t]  
    \centering
    \includegraphics[width=\textwidth]{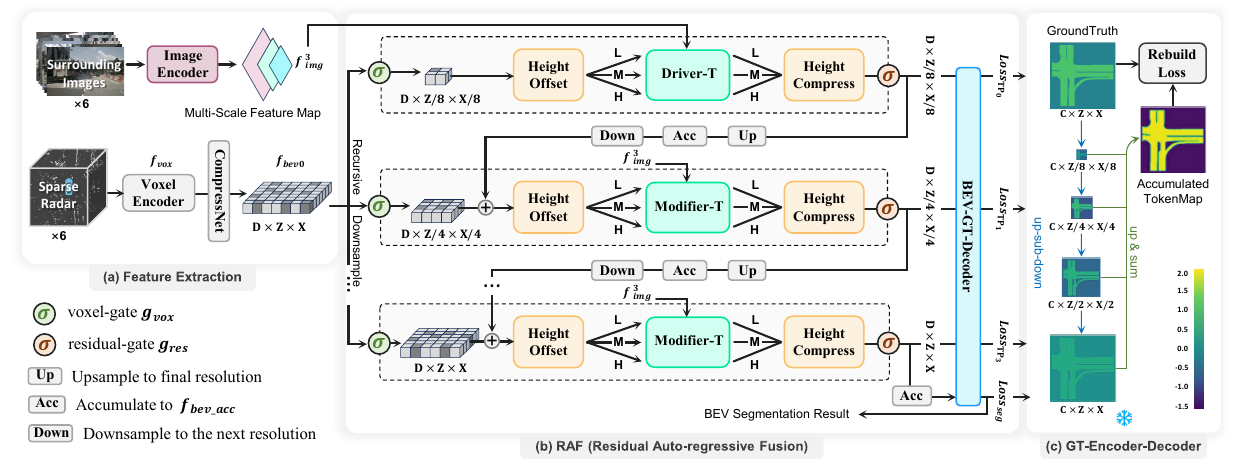}
    \vspace*{-0.4cm}  
    \caption{
Our progressive residual-autoregressive BEV segmentation framework: (a) A dual-branch encoder processes camera and radar inputs. (b) Within the RAF module, the Driver-Transformer generates a low-resolution BEV via cross-modal attention (with dynamic Height-Offset/Compress for ground features), and is followed by the Modifier-Transformer, which predicts multi-scale residuals through autoregressive refinement, integrating historical outputs and radar features via resolution/channel-wise gates. Each stage computes residual difference losses decoded by the BEV-GT-Decoder, while the final output generates segmentation maps and segmentation loss. (c) A GT-Encoder-Decoder decomposes ground truth into multi-resolution residuals for hierarchical supervision.
    }
    \label{fig:main_model}
    \vspace*{-0.2cm}  
\end{figure*}

\section{Method}

\subsection{Multi-Scale Ground-Truth Token Maps Decomposition}
\label{ssec:gt_enc_dec}
Unlike prior work that directly predicts full-resolution segmentations, we propose a hierarchical residual learning paradigm via inverse decomposition of the Ground Truth ($GT$). As showing in Fig. \ref{fig:main_model} (c), given a $C\times Z\times X$ annotation $GT$ where $C$  denotes the number of classes, our method progressively decomposes $GT$ into multi-scale token maps ($TPs$). Each $TP$ encodes distinct hierarchical semantics.

{\parskip=3pt
\noindent\textit{GT Decomposition:} 
Inspired by the multi-scale residual discretization in RQ-VAE~\cite{rqvae}, we design an up-sub-down decomposition process (Alg. \ref{alg:gt_decomp}) that progressively decomposes the original $GT$ ($Z\times X$) into multi-scale token maps $TP_i$. The process initializes the $GT$ as residual $R_1$, then at the $i$-th level of the $N$-level hierarchy, downsampling $R_i$ to  $TP_i$ with resolution $Z/2^{N-i}\times X/2^{N-i}$  while updating $R_{i+1}$ through residual subtraction, until finally outputting the last residual $R_N$ as full-resolution $TP_N$.
To ensure numerical stability and suppress noise propagation, we propose a cascaded residual update mechanism combining dynamic gating $\sigma(\theta)$ and $tanh(\cdot)$ nonlinearity:

\begin{equation}
\label{res_upd}
R_{i+1} = R_i - \sigma(\theta^{(c)}_i)\odot \text{UP}(\text{tanh}(\text{Down}(R_i)))
\end{equation}

$tanh(\cdot)$ confines each token map $TP_i$ within $(-1,1)$, enabling stable residual fluctuations around binary $GT$ values while preventing feature explosion that could impair BEV-GT-Decoder convergence. The resolution- and channel-wise gating ($\sigma(\theta) \in (0,1)^{(N\times C)}$) adaptively modulates cross-scale residual retention, maintaining balanced feature transitions across BEV hierarchies through learnable per-channel attenuation. From the update rule in Eq. \ref{res_upd}, the element-wise update can be expressed as (upsampling omitted from description for clarity due to its negligible effect on the numerical range):

\begin{equation}
\setlength{\abovedisplayskip}{-5pt}
\label{elem_upd}
|u_{ijk}| = |\sigma(\theta_{ij}) \cdot \tanh(\text{Down}(R_{ijk}))| \leq |\sigma(\theta_{ij})|
\end{equation}

where j denotes the class index, and k the pixel index. The corresponding $\ell_2-norm$ of the update satisfies:

\begin{equation}
\label{norm2}
\|U_{ij}\|_2 = \sqrt{\sum_k u_{ijk}^2} \leq \sqrt{d} \cdot \sigma(\theta_{ij})
\end{equation}

where d is the number of elements. 
Further, we define the relative update rate at level i as:
\begin{equation}
\label{upd_rate}
\gamma_i := \frac{\| R_i - R_{i+1}\| }{\| R_i\| } = \frac{\| \sigma_i \odot \tanh(\text{Down}(R_i))\| }{\| R_i\|}
\end{equation}

Leveraging the Lipschitz continuity $(L=1)$ of the hyperbolic tangent function, which satisfies $\|\tanh(\text{Down}(R_i))\|\leq \|\text{Down}(R_i)\|$, we derive the following upper bound:

\begin{equation}
\label{upper_bound}
\gamma_i \leq \sigma^{(i)}_{\max} \cdot \frac{\| \text{Down}(R_i)\| }{\|R_i\|}
\end{equation}

where $\sigma^{(i)}_{\max}$ denotes the maximum value among all channel-wise gating coefficients at the i-th resolution level. This inequality shows that the update in each stage is constrained proportionally to $\sigma^{(i)}_{\max}$, ensuring stable and gradual refinement while mitigating the risk of propagation of noise across scales. Subsequent experiments validate that our gating mechanism ($\sigma(\theta)$ and $tanh(\cdot)$)  ensures stable update dynamics and numerical stability in GT decomposition.

For downsampling, we introduce a hybrid operator combining average pooling with learnable channel-separated convolutions, preserving local geometric structures while dynamically adjusting cross-resolution feature distributions. During upsampling, parameter-free bicubic interpolation maintains geometric consistency between upsampled $TP_i$ and reconstructed $\hat{GT}$, where $\hat{GT}=\sum_i \text{UP}(TP_i)$ (here $\text{UP}(\cdot)$ denoting upsampling to the final resolution) is optimized via Residual Dice Loss (Eq. \ref{seg_loss}) to match BEV segmentation targets.

The pretrained GT-Encoder-Decoder provides fixed multi-scale supervision for the autoregressive BEV network. Offline decomposition during annotation maintains real-time inference without computational overhead.

\begin{algorithm}[t]
\caption{Multi-scale Ground Truth Decomposition}\label{alg:gt_decomp}
{\linespread{1.23}\selectfont
\begin{algorithmic}[1]
\renewcommand{\algorithmicrequire}{\textbf{Input:}}
\renewcommand{\algorithmicensure}{\textbf{Output:}}
\REQUIRE Original mask $GT \in \mathbb{R}^{C \times Z \times X}$, levels $N$
\ENSURE Token maps $\{TP_i\}_{i=1}^N$, reconstructed $\hat{GT}$
\STATE {\textsc{DECOMPOSE}}$(\mathbf{GT}, N)$
\STATE \hspace{0.5cm}$ R_1 \gets \mathbf{GT} $
\STATE \hspace{0.5cm}$ \hat{\mathbf{GT}} \gets \mathbf{0}$
\STATE \hspace{0.5cm}\textbf{for} $ i = 1 $ \textbf{to} $ N-1 $ \textbf{do}
\STATE \hspace{0.9cm}$ \mathbf{TP}_i \gets \tanh(\textsc{AvgPoolConv}(R_i, (\frac{Z}{2^{N-i}},\frac{X}{2^{N-i}}))$ 
\STATE \hspace{0.9cm}$ \hat{\mathbf{TP}}_i \gets \textsc{Bicubic}(\mathbf{TP}_i, (Z,X)) $ 
\STATE \hspace{0.9cm}$ R_{i+1} \gets R_i - \sigma(\theta_{i}^{(C)})\odot\hat{\mathbf{TP}}_i $ 
\STATE \hspace{0.9cm}$ \hat{\mathbf{GT}} \gets \hat{\mathbf{GT}} + \hat{\mathbf{TP}}_i $ 
\STATE \hspace{0.5cm}\textbf{end for}
\STATE \hspace{0.5cm}$ \mathbf{TP}_N \gets \tanh(R_N) $ 
\STATE \hspace{0.5cm}$ \hat{\mathbf{GT}} \gets \hat{\mathbf{GT}} + \mathbf{TP}_N $
\STATE \hspace{0.5cm}\textbf{return} $ \{\mathbf{TP}_i\}_{i=1}^N, \hat{\mathbf{GT}} $
\end{algorithmic}
}
\label{alg:gt_decomp}
\end{algorithm}

\subsection{MultiModal Data Encoding}

Our framework processes two complementary modalities: semantically rich but depth-lacking camera images and spatially precise but sparse radar point clouds. Prior to fusion, both modalities are encoded independently. All sensor data are synchronized to the front radar's timestamp, while a cascaded coordinate transformation (sensor$\rightarrow$global$\rightarrow$ego) spatially aligns six sweep (0.07s/frame) multi-radar point clouds into a unified ego coordinate frame.

{\parskip=3pt
\noindent\textit{Images Encoding:} 
The image encoder is built upon the ResNet-101~\cite{resnet} architecture, which effectively mitigates the vanishing gradient problem in deep networks through residual connections. Specifically, 6 synchronized images $(H\times W)$ are fed into the first three feature sub-layers , generating multi-scale feature maps. To standardize the channel dimensions, each sub-layer output is followed by a channel compression convolution ($1\times 1$). The final outputs consist of three feature maps with spatial resolutions of $1/4, 1/8, \text{ and } 1/16$ of the original image size, denoted as $f^3_{img}$.

{\parskip=3pt
\noindent\textit{Radar Encoding:}
We employ a voxel-based method to encode point clouds into a BEV-aligned representation. A 3D voxel grid $\mathcal{V}\in \mathbb{R}^{Z\times Y \times X}$ is defined in the front-view camera coordinate system, with synchronized points discretized into voxels, each normalized to 10 points via random dropout or zero-padding. Building on VFE~\cite{vfe}, we enhance it with dual-path pooling: each 6-dim point feature $(x, y, z, v_x, v_y, rcs)$ is first embedded into a D-dim space. Max-pooling then extracts salient local features $p_{max}$, while attention-pooling aggregates contextual features $p_{attn}$. As shown in Fig. \ref{fig:vfe}, point-wise features are concatenated as $[p; p_{max}; p_{attn}] \in \mathbb{R}^{3D}$ and compressed back to D-dim via an MLP. This encoding is cascaded in two stages, followed by voxel-wise max-pooling to generate a compact representation $f_{vox} \in \mathbb{R}^{D \times Z \times Y \times X}$. 

\begin{figure}[h]  
    \centering      
    \includegraphics[width=\columnwidth]{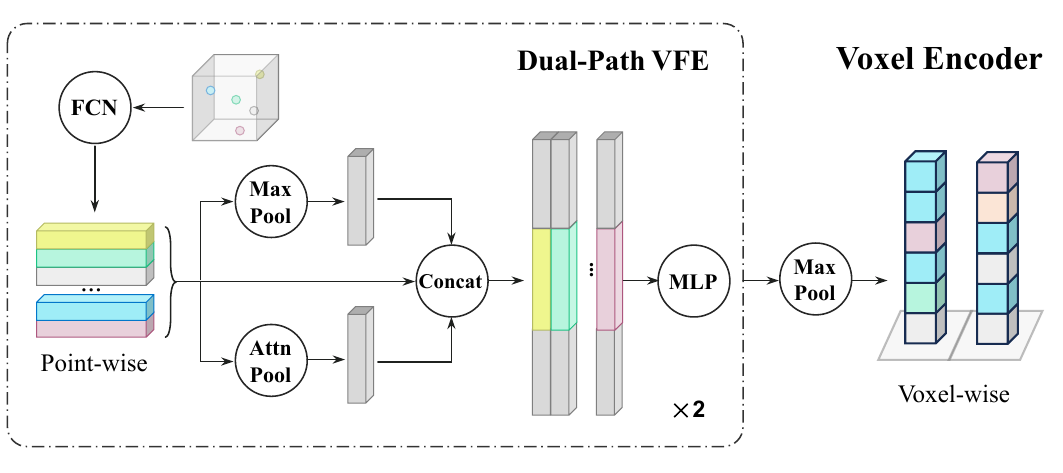} 
    \vspace*{-.6cm}
    \caption{Voxel feature extraction: we normalize each voxel to $10$ points, then extract $C\times 10$ features via point-wise encoding. Apply parallel max/attention-pooling, concatenate with original features $(3C\times 10)$, and compress to C channels via MLP. Repeat twice, then max-pool for final voxel features.}   
    \label{fig:vfe} 
\end{figure}

A CompressNet (Fig. \ref{fig:compressnet}) applies a two stage 3D convolution to downsample the height (Y) dimension. The compressed features are reorganized by merging the height and channel dimensions, and finally refined through a 2D convolution to produce the BEV feature $f_{bev0} \in\mathbb{R}^{(D,Z,X)}$.

\begin{figure}[h]  
    \centering      
    \includegraphics[width=\columnwidth]{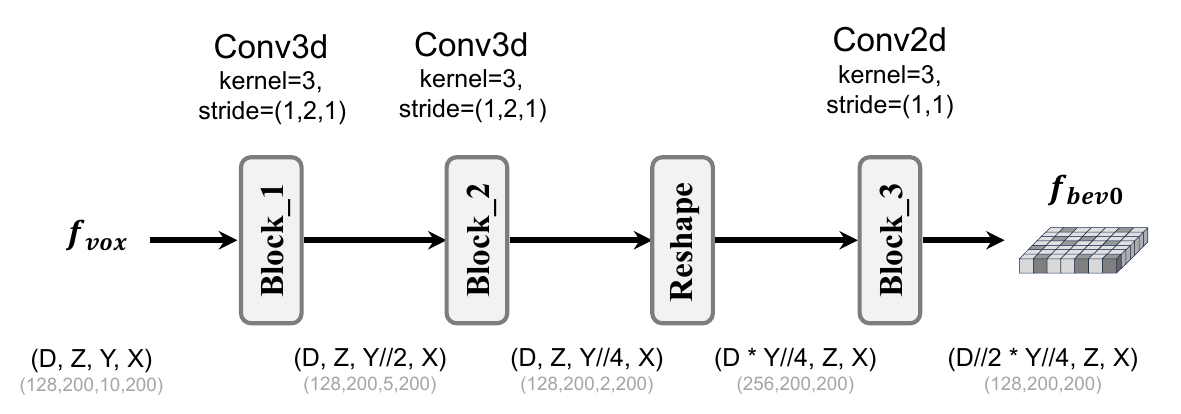} 
    \vspace*{-.6cm}
    \caption{The detailed structure and data flow of compressNet module.} 
    \label{fig:compressnet} 
    \vspace*{-.3cm} 
\end{figure}

\subsection{Ground-Proximity Lifting and Unlifting Process}

In 3D-to-2D perspective interaction, the mapping between imaging plane and voxel grid is established through camera intrinsics. Previous methods~\cite{simplebev,bevcar} use dense ego-centric voxel grids but face drawbacks: high computational cost, irrelevant background noise over the camera (e.g., sky, buildings) and projection deviation away from the ground. As shown in Fig. \ref{fig:lu}(a) and the lifting visualization in Fig. \ref{fig:lu}(b, left), grid regions far from the ground exhibit significant projection errors, while only lower image regions meaningfully interact with the grid in the unlifting result (Fig. \ref{fig:lu}(b, right)).
    
Based on these observations, we propose the ground-proximity projection, constraining the BEV modeling to grid features near the ground. To address ground height uncertainty, we introduce a learnable offset rate $Y_{drift} \in (0,1)$ to adjust the height within a range (±0.6m) relative to a prior $Y_{gr}$ (1m below the camera center):

\begin{equation}
\label{hei_offs}
Y_{new} = Y_{gr}+ofst_{min}+Y_{drift}\cdot (ofst_{max}-ofst_{min}) 
\end{equation}

where $Y_{new}$ is the refined height after adjustment, which better aligns with the actual ground. For multilevel features, three independent $Y_\text{drift}$ values yield initial offsets of -0.277m, 0m, and 0.277m for low-, mid-, and high-level BEV features respectively. To maintain computational efficiency, the sampling coordinates for the ground-proximity projection are computed only once for the highest-resolution voxel grid. Coordinates for lower resolutions are obtained via interpolation.

\begin{figure}[h]  
    \centering      
    \includegraphics[width=\columnwidth]{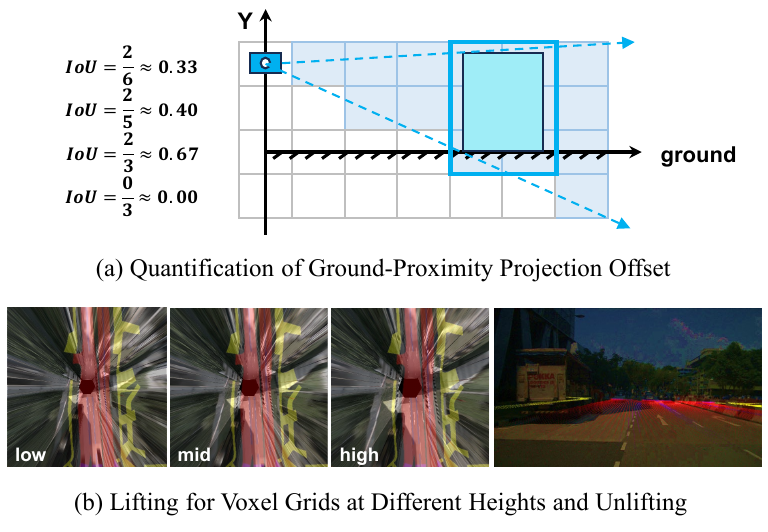} 
    \vspace*{-.6cm}
    \caption{Lifting and unlifting visualization based on camera sensor intrinsics.} 
    \label{fig:lu} 
    \vspace*{-.3cm} 
\end{figure}

\subsection{Residual Autoregressive Fusion}
We propose a progressive Residual Autoregressive Fusion (RAF) module (Fig. \ref{fig:main_model}(b)), which consists of a BEV-initializing drive stage and a multiscale residual-optimizing modify stage. Each stage includes: (i) height-aware positional offsets, (ii) cross-attention Transformer decoders (Driver-T/Modifier-T), and (iii) height compressors. RAF integrates hierarchical voxel features, applies iterative vision-radar cross-modal attention, and refines features via residual accumulation. The output is projected to BEV space by a BEV-GT-Decoder for high-precision segmentation.

{\parskip=3pt
\noindent\textit{Drive Stage:} A convolutional network downsamples $f_{bev0}$  by $1/8$ to produce $f_{bev0} \in \mathbb{R}^{(D,Z/8,X/8)}$ as input of the first-level module. Here, a Height-Offset module enriches \( f_{bev0} \) with low-, mid-, and high-level height information (as defined in Eq. \ref{hei_offs}), followed by the Driver-T interacting with multi-scale image features \( f^{3}_{img} \) (Fig. \ref{fig:transformer}). This module employs deformable attention~\cite{deformableattention} to efficiently model BEV-space locality via dynamic sampling where \( f_{bev0} \) is back-projected onto the image plane for multi-head, multi-scale feature sampling, reducing computation while precisely attending to critical regions. Each grid attends to $\mathcal{P}$ reference points with learnable offsets near the projected locations and fuses features adaptively. The Height-compress module then fuses height-varying features into the lowest-resolution feature \( f'_{bev0} \). In the RAF, \( f'_{bev0} \) is upsampled to the final resolution $(D, Z, X)$ using Bicubic interpolation (consistent with Alg. \ref{alg:gt_decomp}) and accumulated into the zero-initialized $f_{bev\_acc}$. Meanwhile, a learnable convolution downsamples $f_{bev\_acc}$ to $(D, Z/4, X/4)$, generating $f_{bev1}$ for the following residual prediction iteration.  

{\parskip=5pt
\noindent\textit{Modify Stage:} This stage employs a three-level autoregressive refinement process while preserving identical behavior to $GT$ decomposition process. As detailed in Alg. \ref{alg:gt_decomp}, in the $TPs$' generation phase, the $GT$ is decomposed hierarchically through a gated mechanism formulated as:

\begin{equation}
\label{gt_dec}
GT = \sigma(\theta_1)\cdot TP_1 + \sigma(\theta_2)\cdot TP_2 + \sigma(\theta_3)\cdot TP_2 + TP_4  
\end{equation}

where all $TPs$ are processed by $tanh(\cdot)$ activation for value compression. Viewed right-to-left, this represents the BEV segmentation map generation process. Notably, all but the final residual term (which maintains minimal magnitude) are modulated by Sigmoid gates $\sigma(\theta_i)$. To ensure behavioral consistency with $GT$ decomposition, we introduce resolution- and channel-wise learnable residual gating parameters $g_{res}$ at each autoregressive stage (except the final level), dynamically controlling the influence of current $TP_i$ on the accumulated BEV feature $f_{bev\_acc}$. Each level's input features aggregate previous residual outputs:  

\begin{equation}
\label{tp_acc}
f_{bevi}=\text{Down}\Big(\sum_{k=1}^{i-1}\text{Up}(g_{resk}\cdot f'_{bevk})\Big)
\end{equation}

where $\text{Up}(\cdot)$/$\text{Down}(\cdot)$ denote feature resampling. 
Voxel features obtained via multi-scale downsampling are fused with $f_{bevi}$ as input for spatial enhancement. To mitigate conflicts between voxel features and residual accumulations that may introduce high-frequency noise disrupting BEV continuity, we incorporate gating units $g_{vox}$ for voxel features in all cascaded modules inputs (except the drive stage which must use $f_{vox}$ to initialize queries). Simultaneously, a learnable position encoding vector $pos_{enc}\in \mathbb{R}^{Z\times X}$, shared across all-scale voxel features through interpolation sampling, is additionally incorporated.

Modifier-T shares the same architecture as Driver-T but differs in parameterization and sampling strategy. While Driver-T uses independent parameters and a fixed sampling size (P=2) for efficient coarse initialization, Modifier-T employs progressive sampling (P=2,3,4) across resolutions and shares core Transformer parameters across refinement level except for the cross-attention networks (offset prediction and attention weight) , which remain level-specific to capture scale-variant visual cues. This design balances representation capacity and efficiency, enabling Modifier-T to refine details residually while Driver-T focuses on global scene initialization.

\begin{figure}[t!]  
    \centering      
    \includegraphics[width=\columnwidth]{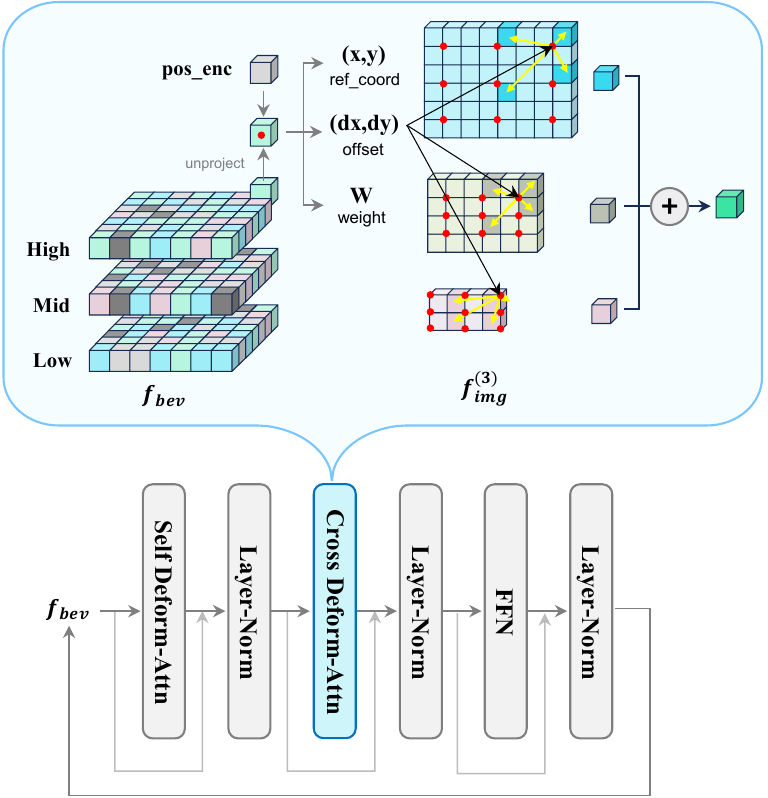} 
    \vspace*{-.6cm}
    \caption{Architecture of Driver/Modifier Transformer decoders. Cascaded decoders process learnable 3-layer $f_{bev}$, where Cross Deformable Attention enables BEV-to-multi-view semantic interaction. Modifier stages maintain independent cross-attention modules while sharing other components.}   
    \label{fig:transformer} 
    \vspace*{-.3cm} 
\end{figure}

\subsection{BEV-GT Decoder and Multi-Scale Supervision}
\label{ssec:decoder_loss}

The RAF module generates multi-scale residual features
\vspace{2pt}
$\{f'_{bevi}\}_{i=1}^4$ decoded into segmentation predictions $\{\hat{TP}_i\}$ via a shared-weight BEV-GT-Decoder (architecturally similar to the segmentation head), matching the GT-Encoder-Decoder's residual outputs $\{TP_i\}$. The accumulated features $f_{bev\_acc}$, formed by aggregating all intermediate results are then decoded by the same head to yield the final prediction $\hat{GT}$.

{\parskip=5pt
\noindent\textit{BEV-GT-Decoder:} A three-stage residual convolutional architecture processes input features $(D,Z_i,X_i)$, performing channel-wise fusion while preserving spatial dimensions. The processed features undergo channel compression through a final $1\times1$ convolution, projecting from dimension $D$ to the target class dimension $C$.

{\parskip=5pt
\noindent\textit{Loss Supervision:}
We design a dual-branch multi-scale supervision mechanism to jointly optimize both residual representations and the final segmentation output:

• Multi-scale Residual Token Map Loss: For intermediate residual predictions $\hat{TP}_i$ at each stage, we employ a configurable feature-level reconstruction loss that supports  $L1 / L2 / Smooth-L1$ norms. To ensure training stability, instead of pixel-wise summation, we compute either channel-wise or spatial-wise differences:  

\begin{equation}
\label{res_loss}
\mathcal{L}_{TP_i}=
\begin{cases} 
\frac{1}{ZX}\sum_{z,x}\|TP_i - \hat{TP}_i\|_p & \text{(Spatial-wise)} \\ 
\frac{1}{BC}\sum_{b,c}\|TP_i - \hat{TP}_i\|_p & \text{(Channel-wise)} 
\end{cases}
\end{equation}

• Adaptive Segmentation Dice Loss: The final BEV segmentation output is supervised using a class-adaptive weighted Dice loss:

\begin{equation}
\label{seg_loss}
\mathcal{L}_{seg} = \frac{1}{C}\sum_{c=1}^C w_c\left(1 - \frac{2\sum p_c \cdot g_c + \epsilon}{\sum p_c + \sum g_c + \epsilon}\right) 
\end{equation}
where $p_c = \sigma(\hat{y}_c)$ is the sigmoid-normalized prediction logits 
probability for class $c$, $g_c$ denotes the original binary value of $GT$, 
and $w_c = \frac{1-f_c}{\frac{1}{C}\sum(1-f_c)}$ is the adaptive weight for class $c$, with $f_c$
being class frequency. The smoothing term $\epsilon=10^{-5}$ ensures numerical stability.
}

\section{Experiments}

In this section, we will elaborate on the experimental setup, the comparison between RESAR-BEV and other single-stage segmentation BEV models, as well as the impact of key modules and hyperparameters in the RESAR-BEV model on both training and final inference performance.

\begin{table*}[t]
\footnotesize
\centering
\caption{Performance Comparison on nuScenes Dataset}
\label{tab:baselines}
\renewcommand{\arraystretch}{1.2} 
\begin{tabular}{@{}l ccc ccc cc c@{}}
\hline
\multirow{2}{*}{\textbf{Method}} & \multirow{2}{*}{\textbf{Mod.}} & \multirow{2}{*}{\textbf{Res.}} &\multirow{2}{*}{\textbf{Backbone}} & 
\multicolumn{3}{c}{\textbf{Segmentation IoU (\%)) $\uparrow$}} & \multirow{2}{*}{$\Delta$\textbf{mIoU\% $\uparrow$}} & \multirow{2}{*}{\textbf{Param.}} & \multirow{2}{*}{\textbf{FPS $\uparrow$}} \\
\cline{5-7}
 & & & &\textbf{Drivable Area} & \textbf{Vehicle} & \textbf{Lane Divider} & & \\
\hline
\multicolumn{10}{@{}l}{\textbf{\textit{Camera-only}}} \\

LSS~\cite{lss}$^\dagger$               & C   & 128$\times$352  & EffNetB0   & 72.94 \textcolor{green}{($+$\phantom{0}4.11)}  & 32.07 \textcolor{green}{($+$14.77)}       & 19.96 \textcolor{green}{($+$15.33)}& \textcolor{green}{ $+$\phantom{0}9.44} / \textcolor{green}{$+$11.40}           & 14.3M & 25.0 \\

CVT~\cite{cvt}$^\dagger$               & C   & 224$\times$448  & EffNetB4   & 74.30 \textcolor{green}{($+$\phantom{0}6.70)} & 36.00 \textcolor{green}{($+$18.98)}       &  --                              & \textcolor{green}{$+$12.84} / --\phantom{00000}                   & 4.3M & \textbf{34.0} \\

BEVFormer~\cite{bevformer}$^\dagger$   & C   & 900$\times$1600 & Res101       & 77.50 \textcolor{green}{($+$\phantom{0}7.72)}  & 46.70 \textcolor{green}{($+$11.51)}       & 23.90 \textcolor{green}{($+$22.37)}& \textcolor{green}{ $+$\phantom{0}9.62}  / \textcolor{green}{$+$13.87}           & 75.0M & 1.7 \\

BEVFormer-S~\cite{bevformer}$^\dagger$ & C   & 900$\times$1600 & Res101       & 80.70 \textcolor{green}{($+$\phantom{0}4.52)}  & 43.20 \textcolor{green}{($+$15.01)}       & 21.30 \textcolor{green}{($+$24.97)}& \textcolor{green}{ $+$\phantom{0}9.77}  / \textcolor{green}{$+$14.83}           & 68.7M & -- \\

CGLSCL~\cite{cglscl}$^\dagger$         & C   & 900$\times$1600 & Res50        & 80.40 \textcolor{green}{($+$\phantom{0}4.28)}  & 41.60 \textcolor{green}{($+$16.19)}       & --                               & \textcolor{green}{$+$10.24} / --\phantom{00000}                   & \textbf{2.7M}  & --\\

DiffBEV ~\cite{diffbev}$^\dagger$      & C   & 800$\times$600  & SwinT       & 65.40 \textcolor{green}{($+$18.60)} & 38.90 \textcolor{green}{($+$18.10)}       & --                               & \textcolor{green}{$+$18.35} / --\phantom{00000}                   & 78.8M  & 11.7\\

GENBEV ~\cite{genbev}$^\dagger$        & C   & 800$\times$600  & Res50        & 74.10 \textcolor{green}{($+$\phantom{0}9.21)}  & 44.60 \textcolor{green}{($+$11.72)}       & --                               & \textcolor{green}{$+$10.47} / --\phantom{00000}                   & 86.4M  & 7.0\\

\hline
\multicolumn{10}{@{}l}{\textbf{\textit{Camera+Radar}}} \\

BEVGuide~\cite{bevguide}$^\dagger$     & C+R & 224$\times$480  & EffNet      & 76.70 \textcolor{green}{($+$\phantom{0}5.71)}  & 59.20 \textcolor{red}{($-$\phantom{0}3.62)} & 44.20 \textcolor{red}{($-$\phantom{0}0.90)}   & \textcolor{green}{ $+$\phantom{0}1.05} / \textcolor{green}{$+$\phantom{0}0.40}            & -- & 24.0 \\
Simple-BEV~\cite{simplebev}$^{\ddagger}$            & C+R & 448$\times$672  & Res101       &  78.26 \textcolor{green}{($+$\phantom{0}5.27)}   & 52.64 \textcolor{green}{($+$\phantom{0}4.23)}    &  37.81 \textcolor{green}{($+$\phantom{0}6.62)} &  \textcolor{green}{$+$\phantom{0}4.75} / \textcolor{green}{$+$\phantom{0}5.37}  & 42.2M & 7.6 \\
CRN~\cite{crn}$^{\ddagger}$                         & C+R & 448$\times$672  & Res101       & 81.34 \textcolor{green}{($+$\phantom{0}2.19)}   & 54.80 \textcolor{green}{($+$\phantom{0}2.07)}    &  39.69 \textcolor{green}{($+$\phantom{0}4.74)} & \textcolor{green}{$+$\phantom{0}2.13} / \textcolor{green}{$+$\phantom{0}3.00}   & 117.7M  & 25.0 \\
BEVCar~\cite{bevcar}$^{\ddagger}$                   & C+R & 448$\times$672  & Res101       & 79.40 \textcolor{green}{($+$\phantom{0}4.13)}   & 54.17 \textcolor{green}{($+$\phantom{0}2.70)}  & 43.07 \textcolor{green}{($+$\phantom{0}1.36)} & \textcolor{green}{$+$\phantom{0}3.41} / \textcolor{green}{$+$\phantom{0}2.73}      & 95.5M & 2.6 \\
\hline
\multicolumn{10}{@{}l}{\textbf{\textit{Our Approach}}} \\

RESAR-Camera                 & C   & 448$\times$672 & Res101   & 76.88 & 46.60 & 40.20 & -- / -- & 30.8M & 17.1 \\
RESAR-E2E             & C+R & 448$\times$672 & Res101   & 77.10 & 52.90 & 41.50 &  -- / -- & 31.0M & 15.5 \\ 
RESAR-Standard                 & C+R & 448$\times$672 & Res101   & 83.53 & 56.87 & 44.43 & -- / -- & 31.9M & 14.6  \\

\hline
\end{tabular}

\vspace{0.6em}
\footnotesize
\raggedright
\textbf{Abbr. :}    C: Camera;\, R: Radar;\, $\dagger$: official results;\, ${\ddagger}$: rerun under our config;\, ($\pm$x): difference in IoU under identical experimental settings (RESAR-Standard$-$baseline);\, $\textbf{Bold}$: Best performance in each category;\, $\uparrow$/$\downarrow$: higher/lower are better.\\
\vspace{0.5em}
We evaluate RESAR-BEV against camera-only and camera-radar unidirectional end-to-end baseline models on the nuScenes validation set. To account for variations across approaches, we assess three key autonomous driving segmentation tasks: Drivable Area, Vehicle, and Lane Divider, using both individual IoU and mIoU increment (left: mean of first two categories; right: mean of all three categories). We also compare model parameters and inference speed to evaluate real-time performance. Ablation studies validate RESAR's camera-only and end-to-end configurations.
\end{table*}

\subsection{Dataset and Evaluation Protocol}

Our model is evaluated on the nuScenes~\cite{nuscenes} v1.0 full dataset, comprising 850 driving scenes (700 training/150 validation) with 34149 samples captured in Boston and Singapore. The dataset provides synchronized multi-sensor data including six 360° Cameras, LiDAR, and Radar. Following the setting in BEV-Car, we focus on cost-effective but challenging sensor configurations using only Camera and Radar, targeting seven critical BEV segmentation categories: drivable area, pedestrian crossing, walkway, stop line, road divider, lane divider, and vehicle. For comprehensive evaluation, we employ multiple metrics including mIoU (mean intersection over union) for segmentation accuracy, per-class IoU for detailed category analysis, FPS (frames per second) for inference efficiency, parameter count for model complexity, as well as range interval evaluation to assess distance-dependent performance degradation, with additional testing under adverse weather conditions to examine robustness.

\subsection{Experimental Settings}

{\parskip=3pt
\noindent\textit{BEV Scene Configuration:} We established a unified spatial coordinate system centered on the front-facing camera at a prior height of 1m above the ground. To represent radar point cloud features, we constructed a voxel grid covering a range of $\pm 50$m along the driving direction (Z-axis) and lateral direction (X-axis), with a resolution of $200\times 200$ ($0.5$m / pixel). In the vertical dimension (Y-axis), the grid spans $\pm 5$m at a resolution of 0.2m (50 voxel bins). The vertical coordinates were scaled by a factor of 1/5 (10m$\rightarrow$2m range) via voxel resizing while preserving the spatial aspect ratio. This operation reduces the number of vertical bins from 50 to 10, improving computational efficiency while retaining essential spatial information. The three BEV feature grids were initialized near the prior ground height at -0.277m, 0m, and 0.277m, with learnable offsets constrained within a range of ±0.6 meters. Input images from all six cameras were uniformly resized and cropped to a resolution of $448\times 672$, and temporally aligned six-frame point clouds were employed to compensate for the sparsity of the radar data.

}

{\parskip=3pt
\noindent\textit{Model Structure:} All embedding layers are standardized to 128 dimensions, while compressed voxel features are integrated with learnable positional encoding. Both Driver-T and Modifier-T employ two-stage Transformer decoders, though Modifier-T's cross-attention modules operate independently across resolutions. A weighted balance is applied between four-stage resolution losses (weights $2.0 / 3.0 / 4.0 / 5.0$) and the segmentation loss ($10.0$). Both residual and radar gating use channel/resolution-wise learnable vectors.
}

\subsection{Benchmark Performance Comparison} 

{\parskip=3pt
We selected three common categories for a multi-dimensional comparison with existing baselines (LSS~\cite{lss}, CVT~\cite{cvt}, BEVFormer~\cite{bevformer}, CGLSCL~\cite{cglscl}, DiffBEV~\cite{diffbev}, GENBEV~\cite{genbev}, Simple-BEV~\cite{simplebev}, CRN~\cite{crn}, BEVGuide~\cite{bevguide}, and our primary baseline BEVCar~\cite{bevcar}).
To ensure a fair comparison, we adopt two alignment protocols: adapting RESAR-BEV to the baseline’s original configuration or re-implementing the baselines under our setup, the latter encompassing image resolution, backbone, BEV grids, and output categories. As shown in Table~\ref{tab:baselines}, RESAR-BEV surpasses all baseline methods in terms of average precision. For the combined metric of Drivable Area and Vehicle segmentation, our model achieves a $+1.05\%$ improvement over the strongest baseline, while maintaining a $+0.40\%$ lead in the three-category combined metric.
Notably, our full model contains only $33.4\%$ of the parameters of BEVCar while achieving $5.62\times$ faster FPS, also surpassing both Simple-BEV and the temporal fusion-based BEVFormer. Ablation studies reveal that the end-to-end variant without residual supervision outperforms the camera-only version, validating the importance of Radar depth information in challenging scenarios such as nighttime and occlusions. However, the end-to-end model still falls short of our full progressive residual auto-regressive architecture, further confirming the effectiveness of our proposed residual supervision mechanism.

\subsection{Robustness Performance} 

{\parskip=3pt
\noindent\textit{Distance:}  Our distance-based robustness analysis evaluates vehicle segmentation performance across varying distances using a $0.5$m/pixel grid centered on the ego vehicle. In Tab. \ref{tab:range}, the CRN model demonstrates superior close-range ($0-20$m) detection by effectively combining Camera and Radar modalities with its dedicated vehicle segmentation head. While our model achieves comparable performance to BEV-Car in mid-range ($20-35$m) scenarios, It achieves a long-range ($35-50$m) detection performance of $40.8\%$, significantly outperforming all baselines. The overall performance improvement of $2.1\%$ across all ranges ($0-50$m) can be attributed to our progressive residual refinement strategy, which first captures coarse vehicle locations before performing fine-grained residual corrections, combined with the robust depth estimation provided by Radar fusion, particularly beneficial in challenging conditions like low-light environments or occluded scenarios.

To further evaluate the range extension, we conducted a scalability analysis by expanding the physical coverage from 100×100 m to 200×200 m while maintaining the original BEV grid size of 200×200 pixels. This reduces the ability to discern fine-grained structures such as lane dividers and vehicles (Fig. \ref{fig:long_range}). Maintaining the original resolution of 0.5m/pixel would quadruple the memory load in the core modules (Fig. \ref{fig:complex}). Therefore, the original configuration was selected to balance efficiency and performance.
\begin{table}[h]
\renewcommand\arraystretch{1.1}
\caption{Perception Range Performance Comparison \\ for Vehicle Segmentation}
\label{tab:range}
\setlength{\tabcolsep}{7pt}
\begin{tabular}{lccccc}
\hline
\multirow{2}{*}{\textbf{Method}} & 
\multirow{2}{*}{\textbf{Modality}} & 
\multirow{2}{*}{\textbf{0-50m}} & 
\multicolumn{3}{c}{\textbf{Range Intervals (m)}} \\
\cline{4-6}
 & & & \textbf{0-20m} & \textbf{20-35m} & \textbf{35-50m} \\
\hline

Simple-BEV$^{\ddagger}$ & C+R & 52.6 & 71.4 & 52.1 & 34.4 \\
CRN$^{\ddagger}$ & C+R & 54.8 & \textbf{81.4} & 47.1 & 35.9 \\
BEVCar$^{\ddagger}$ & C+R & 54.2 & 73.2 & 50.8 & 38.5 \\
\hline
RESAR & C & 46.6 & 69.2 & 44.9 & 25.5 \\
RESAR-E2E & C+R & 52.9 & 74.0 & 47.8 & 36.8 \\
\textbf{RESAR} & C+R & \textbf{56.9} & 77.6 & \textbf{52.2} & \textbf{40.8} \\
\hline
\end{tabular}

\vspace{0.6em}
\footnotesize
\raggedright
\textbf{Abbr.:} 
${\ddagger}$: rerun under our config; All values represent mIoU (\%).
\end{table}

\begin{figure}[h]  
    \centering      
    \includegraphics[width=\columnwidth]{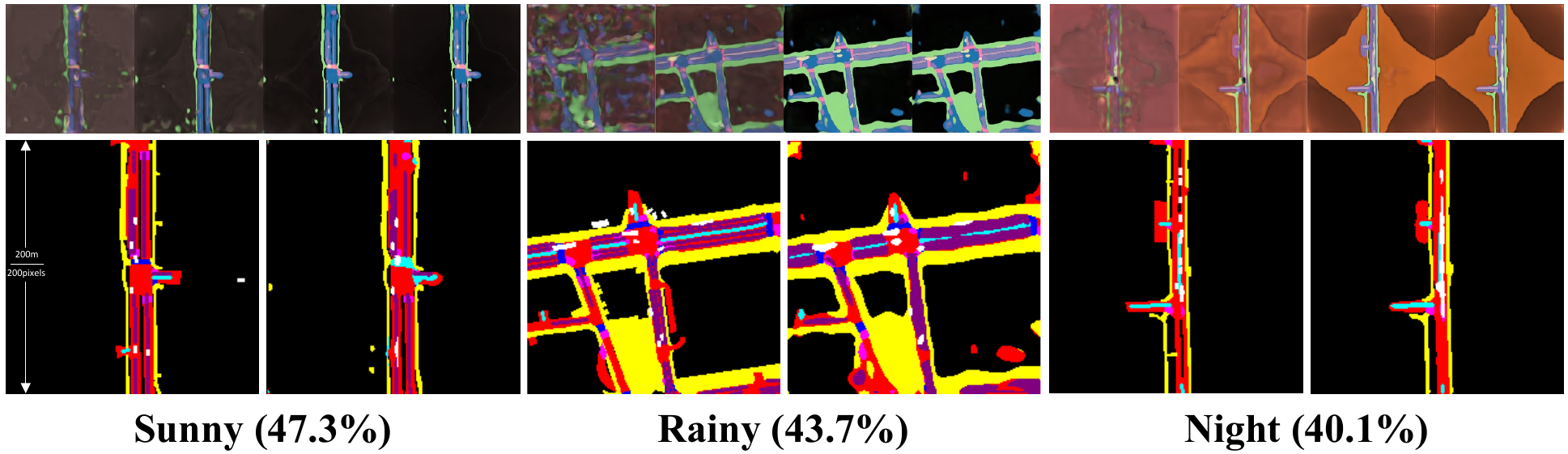} 
    \vspace*{-.6cm}
    \caption{Experimental results of maintaining $200 \times 200$ BEV resolution while expanding physical perception range to $200m \times 200m$.}   
    \label{fig:long_range} 
\end{figure}

\begin{figure}[h]  
    \centering      
    \includegraphics[width=\columnwidth]{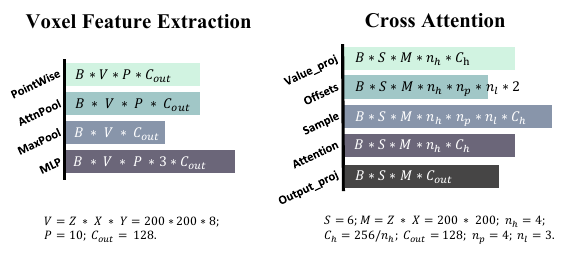} 
    \vspace*{-.6cm}
    \caption{Visualization of spatial complexity proportion in cross-attention module and radar encoding module.}   
    \label{fig:complex} 
\end{figure}

\noindent\textit{Environment:} Our environment-based robustness analysis reveals complementary camera-radar limitations. Using the validation split from ~\cite{bevcar}, we benchmarked our full progressive refinement model, its end-to-end variant, and BEVCar across three critical driving scenarios: sunny (optimal), rainy (moderate), and dark (severe). In Tab. \ref{tab:whether}, our key findings: (1) All models demonstrate improved drivable area and vehicle segmentation performance in rainy conditions, attributed to reduced vehicular occlusion and more comprehensive Camera coverage compensating for Radar deficiencies, while lane segmentation accuracy declines due to water surface reflections. (2) Nighttime degradation primarily stems from cameras losing reliable visual cues, while radar maintains consistent performance despite its inherent sparsity. The comprehensive retesting confirms our full model's robustness ($54.0\%$ mIoU), achieving a $3.23\%$ advantage over BEVCar ($50.77\%$) across all categories and conditions.

\begin{table}[h]
\renewcommand\arraystretch{1.1}
\caption{Performance Comparison Under \\ Different Weather Conditions}
\label{tab:whether}
\setlength{\tabcolsep}{7pt}
\begin{tabular}{lccccccc}
\hline
\multirow{2}{*}{\textbf{Method}} & \multicolumn{7}{c}{\textbf{Categories}} \\
\cline{2-8}
 & \textbf{D.A.} & \textbf{P.C.} & \textbf{W.W.} & \textbf{S.L.} & \textbf{R.L.} & \textbf{L.D.} & \textbf{V.H.} \\
\hline

\multicolumn{8}{l}{\textbf{\textit{Sunny Conditions}}} \\
BEVCar$^{\ddagger}$    & 81.0 & \textbf{51.4} & 62.0 & 41.6 & 43.8 & \textbf{45.3} & 54.3 \\
Us-E2E    & 78.4 & 47.3 & 61.2 & 41.0 & 44.5 & 42.1 & 53.2 \\
RESAR     & \textbf{84.5} & 50.7 & \textbf{65.3} & \textbf{43.2} & \textbf{48.5} & 45.9 & \textbf{57.9} \\
\hline

\multicolumn{8}{l}{\textbf{\textit{Rainy Conditions}}} \\
BEVCar$^{\ddagger}$    & 81.4 & \textbf{48.1} & 57.4 & 35.1 & 41.6 & \textbf{44.4} & 55.4 \\
Us-E2E    & 79.6 & 44.7 & 58.6 & 37.8 & 39.9 & 41.5 & 55.3 \\
RESAR     & \textbf{86.9} & 47.5 & \textbf{62.5} & \textbf{40.8} & \textbf{46.7} & 44.1 & \textbf{59.5} \\
\hline

\multicolumn{8}{l}{\textbf{\textit{Night Conditions}}} \\
BEVCar$^{\ddagger}$    & 75.8 & 40.0 & 49.4 & 30.8 & 34.9 & 39.5 & 52.9 \\
Us-E2E    & 73.2 & 40.0 & 48.2 & 32.1 & 33.0 & 40.3 & 49.5 \\
RESAR     & \textbf{79.2} & \textbf{42.5} & \textbf{53.8} & \textbf{38.2} & \textbf{40.5} & \textbf{42.8} & \textbf{53.1} \\
\hline
\end{tabular}

\vspace{0.6em}
\footnotesize
\raggedright
\textbf{Abbr. :} ${\ddagger}$: rerun under our config; D.A.: Drivable Area, P.C.: Pedestrian Crossing, W.W.: Walkway, 
S.L.: Stop Line, R.L.: Road Divider, L.D.: Lane Divider, V.H.: Vehicle; All values represent mIoU (\%).
\end{table}

We further conducted a region masking experiment by removing radar point clouds in vehicle areas to analyze their role in multimodal perception. As shown in Fig. \ref{fig:radar_mask}: 1) In daytime, visual cues maintain partial vehicle detection but with increasing uncertainty at long range (reduced prediction areas); 2) At night, radar masking causes significant miss-detections, proving sparse radar provides essential spatial priors when visuals degrade; 3) Across conditions, radar removal introduces positional misalignment, confirming its critical role in BEV spatial alignment. Results demonstrate that sparse radar data acts as a decisive complementary modality under low-light conditions, ensuring robustness when vision-alone fails.

\begin{figure}[h]  
    \centering      
    \includegraphics[width=\columnwidth]{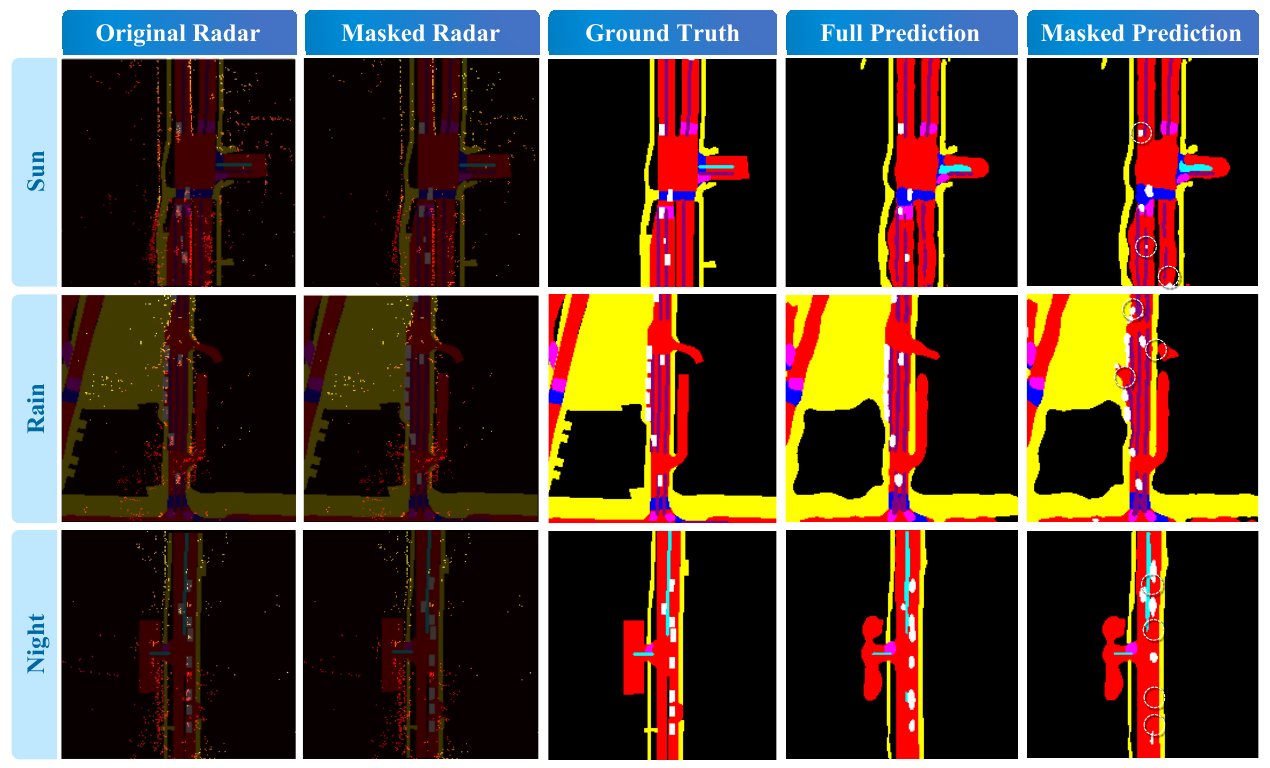} 
    \vspace*{-.6cm}
    \caption{Validating radar's critical role in vehicle perception across day/rain/night conditions via car region masking.}   
    \label{fig:radar_mask} 
\end{figure}

\subsection{Ablation Study}

\begin{figure}[h]  
    \centering      
    \includegraphics[width=\columnwidth]{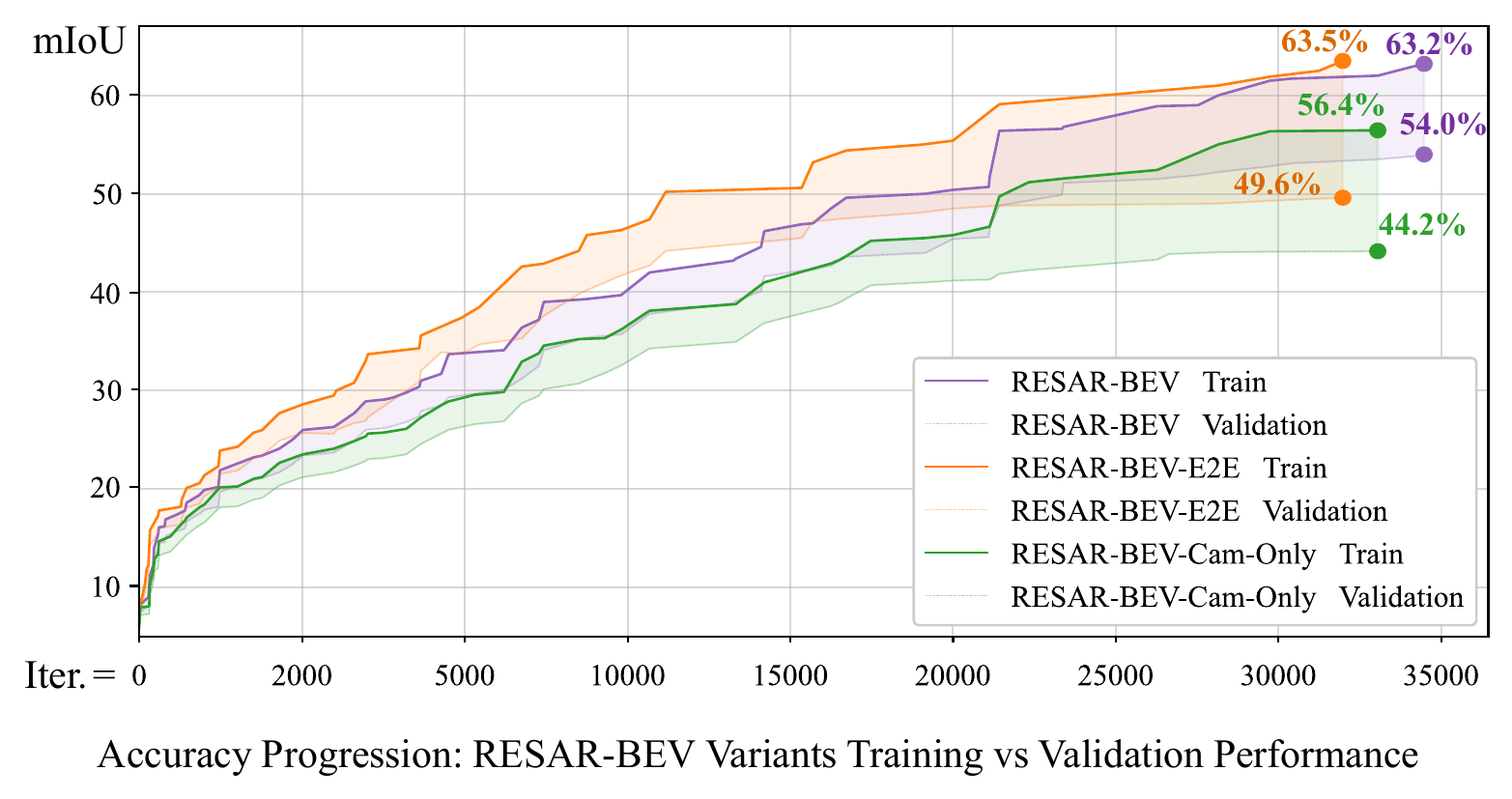} 
    \vspace*{-.6cm}
    \caption{Training and validation mIoU trajectories for complete RESAR-BEV model versus two ablated models across iterations (32 batches/iteration). The semi-transparent bands indicate performance gaps between training and validation sets.}   
    \label{fig:convergence} 
\end{figure}

We conducted comprehensive ablation studies on our RESAR-BEV, evaluating  key modules and critical parameters while comparing inference speed (FPS), model size (parameters), and mean accuracy (mIoU).

{\parskip=3pt
    \noindent\textit{Anti-Overfitting:}  Through a comparative analysis of the complete model, the camera-only model, and an end-to-end model trained without progressive supervision (Fig. \ref{fig:convergence}), our model with Progressive Residual Autoregressive supervision (mIoU $54.0\%$) converges more steadily than end-to-end variant (mIoU $49.6\%$), achieving a higher validation . This effectiveness stems from a multi-faceted anti-overfitting design: 1) Decoupled offline supervision provides stable signals, preventing the model from fitting noise in a direct reconstruction task; 2) Coarse-to-fine progressive learning implements an implicit curriculum, guiding the network from global structures to local details to ensure stable training; 3) Gated autoregressive updating acts as a dynamic regularizer, enforcing controlled, incremental refinements at each step.

{\parskip=3pt
\noindent\textit{Module Ablation:} In Table. \ref{tab:full-ablation}, the camera-only variant eliminates Radar feature extraction but introduces learnable initialization queries $Q_{init}\in\mathcal{R}^{(C,Z/8,X/8)}$ as substitutes, resulting in comparable parameter counts ($-3.4\%$) with $2.5\%$ faster inference yet $9.8\%$ mIoU degradation. Replacing residual learning with direct downsampling causes significant $6.2\%$ accuracy drop. Removing progressive supervision leads to $4.4\%$ performance reduction. Eliminating attention pooling in our proposed VEF decreases accuracy by $5.7\%$.

{\parskip=3pt
\noindent\textit{Parameter Ablation:} In Table. \ref{tab:full-ablation}, disabling both residual gating (for token maps alignment) and voxel gating (for Radar feature modulation) reduces mIoU by $4.7\%$. Fixing three-layer BEV grid offsets decreases accuracy by $5.2\%$. Expanding cascade depth from 2 to 4 levels increases parameters by $3.8\%$ and inference time by $32\%$, yielding marginal $0.4\%$ accuracy gain. Our full model achieves optimal balance with $54.0\%$ mIoU and real-time 14.6 FPS performance.

\begin{table}[h]
\renewcommand\arraystretch{1.1}
\setlength{\tabcolsep}{7.5pt}
\caption{Module \& Hyperparameter Ablation Studies}
\label{tab:full-ablation}
\begin{tabular}{lccccc}
\hline
\multirow{2}{*}{\textbf{Category}} & \multirow{2}{*}{\textbf{Variant}}  & \multicolumn{3}{c}{\textbf{Categories}} \\
\cline{3-5} 
& &\textbf{FPS} & \textbf{Param.} & \textbf{mIoU} \\
\hline

\multicolumn{5}{l}{\textbf{\textit{Module Ablations}}} \\
1 & Camera Only & 17.1 & 30.8M & 44.2 \\
2 & Pyramid or Residual & 15.1 & 31.9M & 47.8 \\
3 & End to End & 15.5 & 31.0M & 49.6 \\
4 & VFE Attention & 15.2 & 31.8M & 51.0 \\
\hline

\multicolumn{5}{l}{\textbf{\textit{Hyperparameter Ablations}}} \\
5 & Voxel, Residual-gate & 14.9 & 31.9M & 49.3 \\
6 & Learnable Height Offset & 14.8 & 31.9M & 48.8 \\
7 & 4 Driver-Modifier Layers & 9.9 & 33.1M & 54.4 \\
\hline

8 & \textbf{Full Model} & 14.6 & 31.9M & 54.0 \\
\hline
\end{tabular}

\vspace{0.4em}
\footnotesize
\raggedright
\textbf{Abbr. :} 
FPS: Frames Per Second, Param.: Parameters (in millions), mIoU: mean Intersection over Union (\%) .
\end{table}

{\parskip=3pt
\noindent\textit{Multi-scale Supervision Weights:} To validate the necessity of multi-scale supervision weights, we compared five weighting configurations (see Fig. \ref{fig:weight_loss}). Results show that while the Ascending configuration performs well, it weakens overall constraints; the Descending configuration converges prematurely to coarse predictions; the Uniform configuration fails to capture multi-scale differences; and the End2End configuration converges slowly. Our proposed weights (2.0, 3.0, 4.0, 5.0) achieve the best mIoU (54.0\%) by balancing multi-scale supervision, confirming their effectiveness.

\begin{figure}[h]  
    \centering      
    \includegraphics[width=\columnwidth]{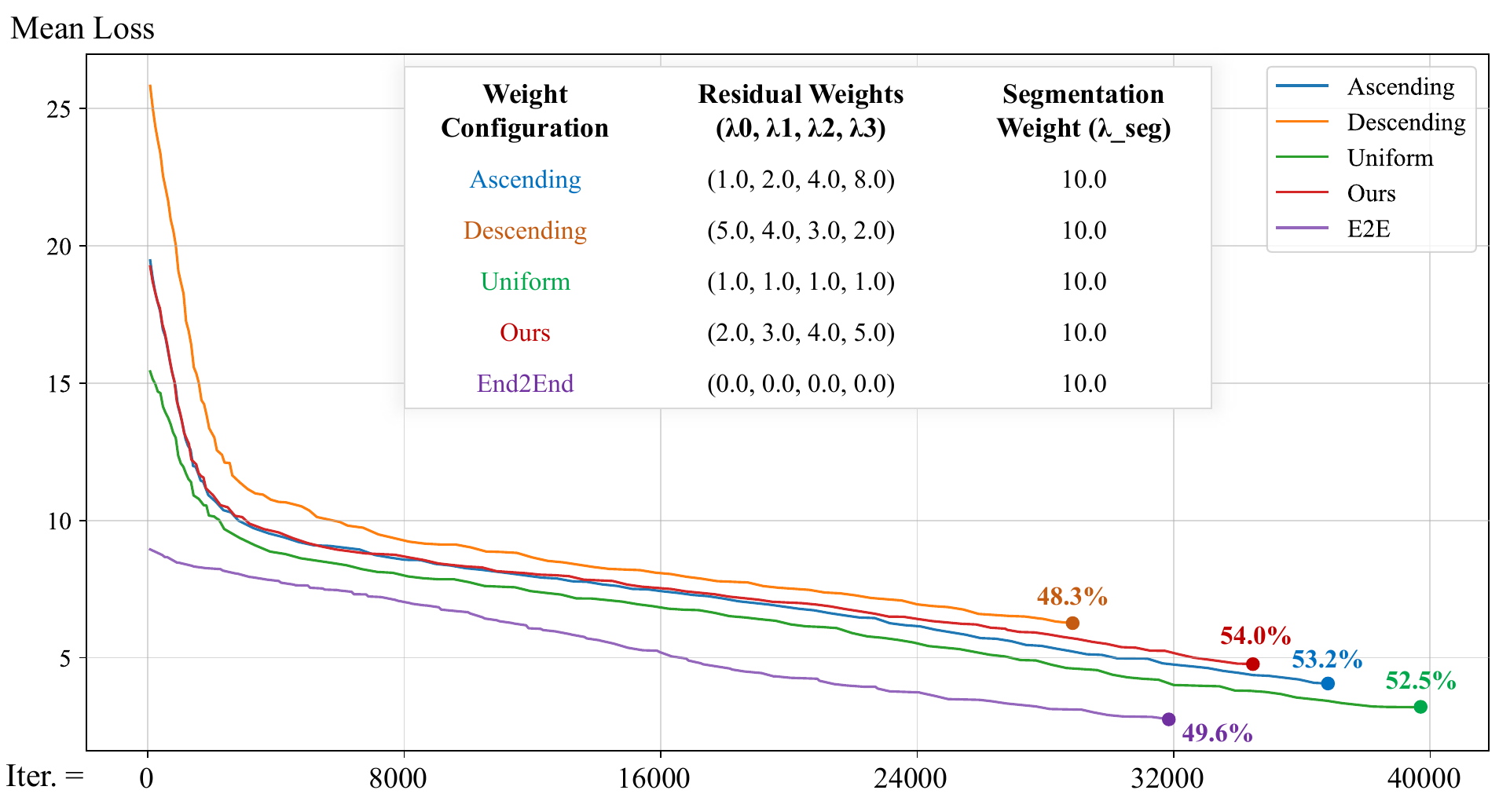} 
    \vspace*{-.6cm}
    \caption{Declining trend of fluctuation-filtered loss and final mIoU across multi-scale supervision weight configurations.}
    \label{fig:weight_loss} 
\end{figure}

{\parskip=3pt
\noindent\textit{Gated Constraint Mechanism:} Based on the $\ell_2-norm$ upper bound in Eq. \ref{norm2}, we compare the empirical mean $\ell_2-norm$ and its theoretical bound under four gating-activation settings (Fig. \ref{fig:gating-exp-vis}). Results show that only the combined use of Sigmoid and Tanh produces stable updates $(<1)$, a uniform distribution, and ensures empirical values remain strictly below the theoretical bound, thereby aiding convergence. Using only Sigmoid leads to large fluctuations $(0.0 \sim 3.0)$ and frequent bound violations, while using only Tanh or no control results in unconstrained updates and noisy signals, impairing training stability. These validate the necessity of joint dynamic gating and Tanh activation in controlling update magnitude and suppressing noise.

\begin{figure*}[h]
    \centering
    \includegraphics[width=\textwidth]{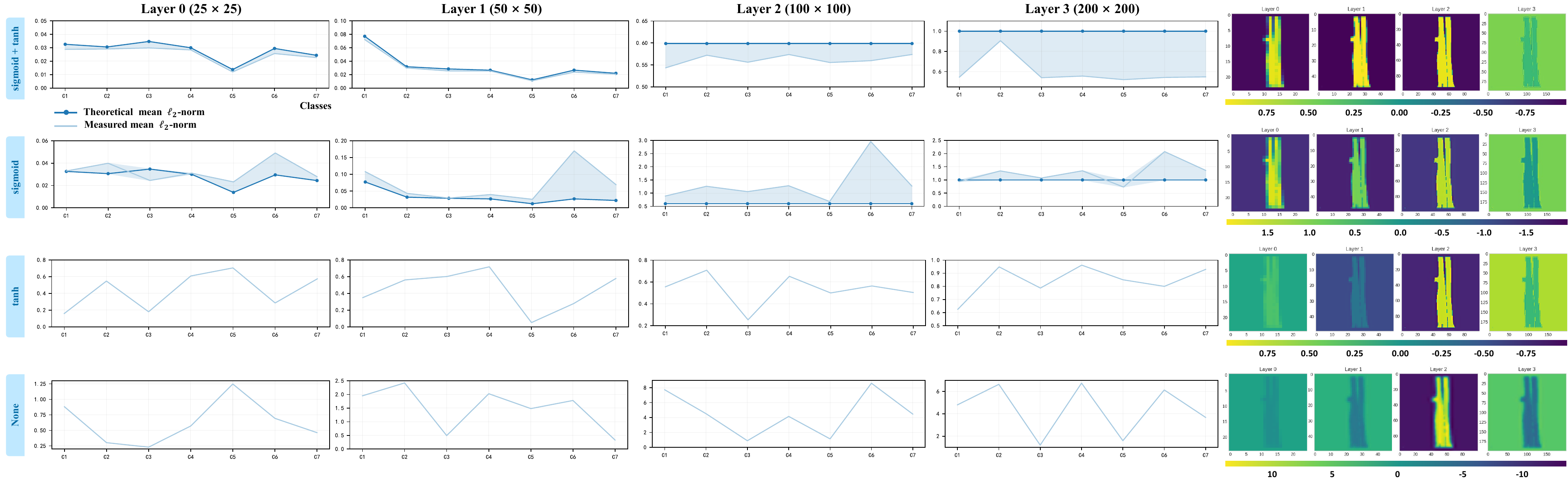} 
    \vspace{-4mm} 
    \footnotesize
    \raggedright
    \caption{
        Experimental validation of the $\ell_2-norm$ constraint effect by dynamic gating and activation functions in the GT Encoding-Decoding process.
    }
    \label{fig:gating-exp-vis}
\end{figure*}

{\parskip=3pt
\noindent\textit{Activation Function:} We ablated activation functions in the GT-Encoder-Decoder module (Fig. \ref{fig:rebuttal_activate}). Tanh consistently outperforms Sigmoid, GELU, ReLU, and Leaky ReLU across all evaluation metrics. It exhibits the smoothest convergence in gating and downsampling parameters, produces the most structurally coherent token maps, achieves the highest reconstruction mIoU, and yields the fastest loss descent. The bounded output and symmetric gradient of Tanh are well-suited for the residual gating and multi-scale learning in our architecture, making it the optimal choice for RESAR-BEV.

\begin{figure*}
    \centering
    \includegraphics[width=\textwidth]{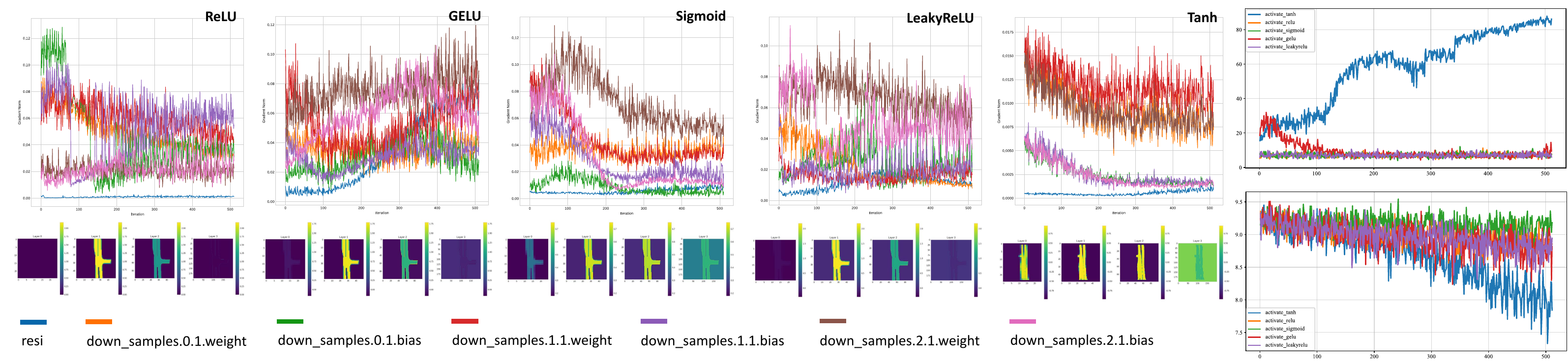} 
    \footnotesize
    \raggedright
    \caption{
        Ablation study analysis of different activation functions on the GT-Encoder-Decoder module: Convergence process of residual gating and downsampling convolution parameters (Top-Left); Visualization of multi-scale Token Maps (Bottom-Left);  Comparison of reconstruction mIoU (Top-Right); Training loss descent curves (Bottom-Right). 
    }
    \label{fig:rebuttal_activate}
\end{figure*}

\subsection{Visualization and Interpretability}
In RESAR-BEV, the multi-scale supervision signals generated by the pre-trained GT-Encoder-Decoder, combined with a stage-wise and objective-specific supervision mechanism, enable precise localization of errors to key generation stages when category prediction inaccuracies occur. This allows the application of differentiated loss weights for targeted optimization, effectively addressing the debugging challenges commonly associated with black-box models.

We present stage-wise residual outputs, their accumulated results, and final threshold-filtered predictions under sunny, rainy, and nighttime conditions (Fig. \ref{fig:vis}). $TP_1$ captures fundamental global semantic and geometric cues of the scene; $TP_2$ and $TP_3$ residuals progressively reveal high-frequency details such as lane markings and object contours; by $TP_4$, the residuals become highly sparse, focusing only on fine-tuning hard-to-predict edges and corners. This coarse-to-fine process aligns with human cognition, confirming distinct semantic abstraction levels across residual stages.

\begin{figure*}[t!]
    \centering
    
    \includegraphics[width=\textwidth]{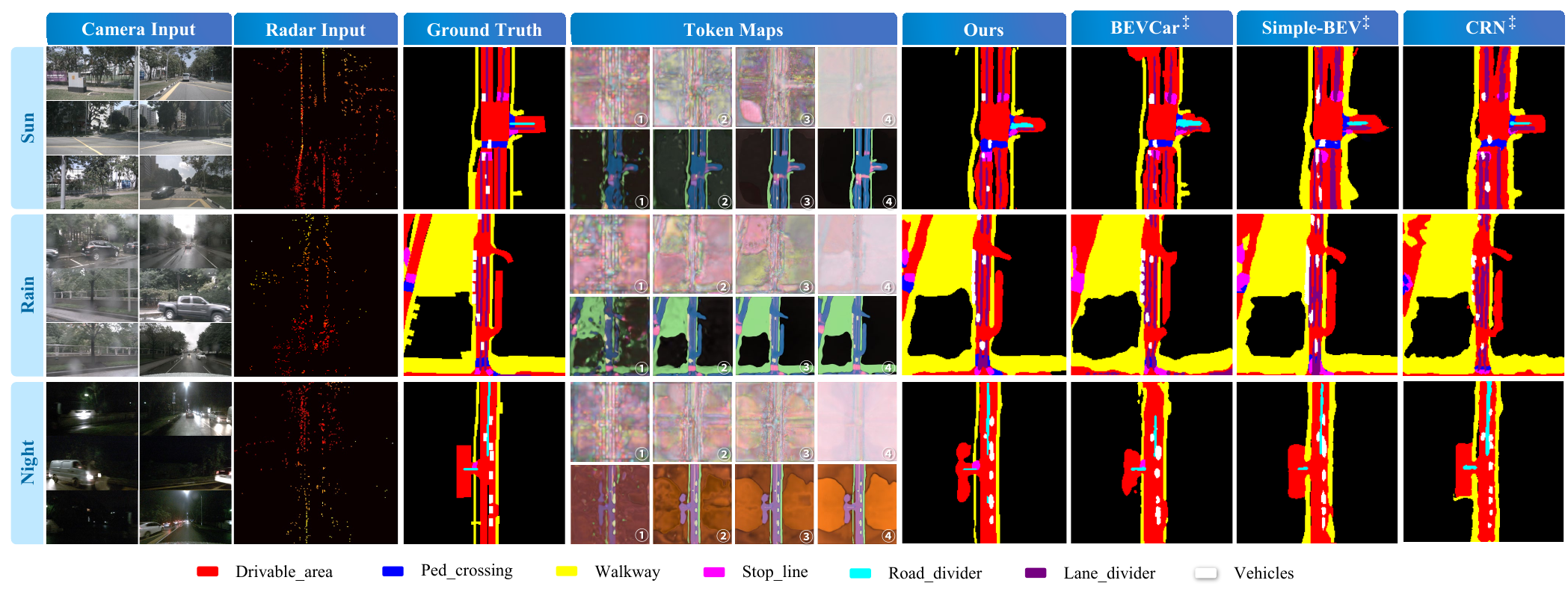} 
    \vspace{-4mm} 
    \caption{
    Progressive multi-modal BEV semantic segmentation via residual auto-regression across diverse environmental conditions. The model integrates synchronized inputs from six surround-view cameras and six consecutive frames of radar point clouds through four-step residual auto-regression shown as stage-wise residuals (top) and accumulated residuals (bottom), outputting seven-class BEV segmentation.
    }
    \label{fig:vis}
\end{figure*}

 We also visualize the attention weights of BEV queries back-projected onto the image plane (Fig. \ref{fig:attnvis}). In low-resolution stages, attention distributions cover broader regions, indicating global topological search. While high-resolution stages becomes more localized. Different attention heads also focus on distinct vertical spatial information, revealing an internally structured and interpretable decision-making process.

\begin{figure}
    \centering
    \includegraphics[width=\columnwidth]{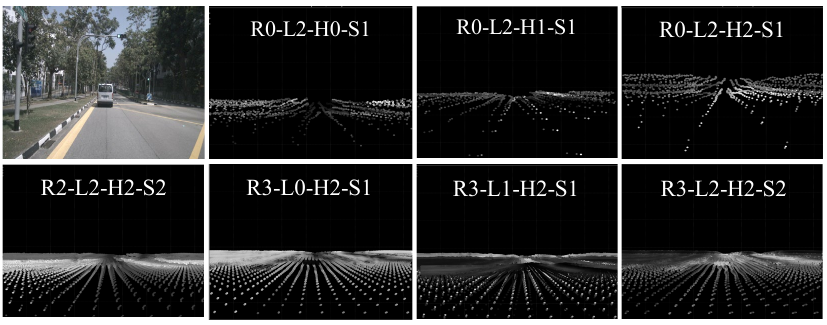}
    \footnotesize
    \raggedright
    \textbf{Abbr. :} 
    \textbf{R}: Residual stage; \textbf{L}: Image feature level; \textbf{H}: Attention head; \\  \hspace*{3.2em}
    \textbf{S}: Driver-T/Modifier-T Decoder layer.
    \caption{
    Visualization of cross-modal attention weights in Driver-T and Modifier-T (image view), showing hierarchical attention patterns across residual modules, feature layers, and decoder levels.
    }
    \label{fig:attnvis}
\end{figure}

This multi-stage evolution aligns perfectly with our residual learning objectives, where early stages establish structural context while later stages focus on local detail recovery.

\section{Conclusion}

RESAR-BEV introduces a progressive residual autoregressive learning framework for BEV segmentation, which decomposes the task into multi-stage residual optimization. By incorporating geometry-guided radar-aided feature querying and an adaptive height offset strategy, it mitigates cross-modal misalignment and error accumulation. Validated on nuScenes, the approach achieves high accuracy, real-time speed, and enhanced robustness in long-range and low-light scenarios. Despite these strengths, RESAR-BEV exhibits certain limitations in complex urban roads, such as multi-lane intersections, rain-induced ground reflections, effective long-range perception and robustness under sensor failure scenarios. Future work will explore temporal-residual architectures with dynamic resolution, enhance robustness to sensor failures via cross-sensor imputation, and improve generalization in challenging urban layouts and reflective surfaces, notably by developing efficient BEV representations that capture high-frequency pixel details for enhanced long-range perception.

\bibliographystyle{IEEEtran}
\bibliography{reference} 

\begin{IEEEbiographynophoto}{Zhiwen Zeng}
is currently pursuing the B.S. degree in computer science and technology at Chongqing University, Chongqing, China. His research interests include autonomous driving and spatial intelligence.
\end{IEEEbiographynophoto}

\begin{IEEEbiographynophoto}{Yunfei Yin} 
 (Member, IEEE) received the B.S. degree in computer science from Peking Univer
sity, the M.S. degree in computer engineering from Guangxi Normal University, and the Ph.D. degree in control science and engineering from Beijing University of Aeronautics and Astronautics. He is currently a Distinguished Research Fellow with the Department of Computer Science, Chongqing University. He mainly engages in research, including artificial intelligence (data mining), system modeling, computer simulation, and unmanned aerial vehicle. In recent years, he has been involved in the National Natural Science Foundation of China, International Large Research Foundation, provincial and ministerial level foundations, and other projects. He has published more than 50 SCI/EI/ISTP-cited refereed articles. He is a reviewer of Journal of Software (Chinese core journal), Artificial Intelligence, and IEEE International Conference on Data Mining (ICDM).
\end{IEEEbiographynophoto}

\begin{IEEEbiographynophoto}{Zheng Yuan}
is currently pursuing the M.S. degree in computer science and technology at Chongqing University, chongqing, China. His research interests include medical image analysis and autonomous driving.
\end{IEEEbiographynophoto}

\begin{IEEEbiographynophoto}{Argho Dey}
is currently pursuing the M.S degree in Computer Technology, at Chongqing University, Chongqing, China. His main research interests include autonomous driving and intelligent sensing.
\end{IEEEbiographynophoto}

\begin{IEEEbiographynophoto}{Xianjian Bao}
received the B.S. degree in computer science from Chongqing University and the M.S. degree in computer engineering from the Maharishi University of Management, Fairfield, IA, USA. He is currently a Distinguished Research Fellow.
\end{IEEEbiographynophoto}

\end{document}